%% file: main.tex
\documentclass[11pt]{article}
\usepackage{ruc-tech-report}
\usepackage{amsmath,amssymb,amsfonts}
\usepackage{booktabs,multirow,array,tabularx}
\usepackage{rotating,pifont,float,subcaption}
\usepackage[export]{adjustbox}
\usepackage{hyperref}
\hypersetup{colorlinks=true,linkcolor=rucred,citecolor=rucred,urlcolor=rucred}
\setkeys{Gin}{max width=\linewidth,max totalheight=0.78\textheight,keepaspectratio}

\newcommand{\authororcid}[2]{{\hypersetup{hidelinks}\href{https://orcid.org/#1}{#2}}}
\providecommand{\Description}[2][]{}
\input{metadata}
\begingroup\fontencoding{T1}\fontfamily{lmr}\selectfont\endgroup
\DeclareFontShape{T1}{lmr}{b}{sc}{<-> ec-qtmb-sc}{}
\DeclareFontShape{T1}{lmr}{bx}{sc}{<->ssub * lmr/b/sc}{}
\hypersetup{pdftitle={What People Choose Not to Say: A Multimodal Benchmark for Fine-Grained Information Manipulation in Conversations},pdfauthor={Yuhan Bian; Zhaokai Liang; Jia Wang; Haoyu Liu; Yu Chen; Jianfeng Qu; Zhixu Li}}
\begin{document}
\input{frontmatter}

\input{body}

\end{document}

%% file: metadata.tex
\newcommand{\papertitle}{What People Choose Not to Say: A Multimodal Benchmark for Fine-Grained Information Manipulation in Conversations}
\newcommand{\paperauthors}{\authororcid{0009-0001-0312-8101}{Yuhan Bian}$^{1,*}$\hspace{0.6em}\authororcid{0009-0004-8744-8998}{Zhaokai Liang}$^{2,*}$\hspace{0.6em}\authororcid{0009-0009-5516-0822}{Jia Wang}$^{1}$\hspace{0.6em}\authororcid{0009-0008-8246-7056}{Haoyu Liu}$^{1}$\\[-0.1em]
\authororcid{0000-0001-8680-1922}{Yu Chen}$^{3,4,\dagger}$\hspace{0.6em}\authororcid{0000-0002-3212-7618}{Jianfeng Qu}$^{3}$\hspace{0.6em}\authororcid{0000-0003-2355-288X}{Zhixu Li}$^{1,3,\dagger}$}
\newcommand{\paperaffiliations}{$^{1}$ School of Information, Renmin University of China, Beijing, China\\
$^{2}$ Faculty of Engineering, The Chinese University of Hong Kong, Hong Kong, China\\
$^{3}$ School of Smart Governance, Renmin University of China, Suzhou, China\\
$^{4}$ State Key Laboratory of Novel Software Technology, Nanjing University, Nanjing, China}
\newcommand{\paperemails}{\href{mailto:xilinghuiye@ruc.edu.cn}{\nolinkurl{xilinghuiye@ruc.edu.cn}}\quad \href{mailto:1155273327@link.cuhk.edu.hk}{\nolinkurl{1155273327@link.cuhk.edu.hk}}\\
\href{mailto:wangjia2025@ruc.edu.cn}{\nolinkurl{wangjia2025@ruc.edu.cn}}\quad \href{mailto:hy.liu25@ruc.edu.cn}{\nolinkurl{hy.liu25@ruc.edu.cn}}\\
\href{mailto:chenyu97@ruc.edu.cn}{\nolinkurl{chenyu97@ruc.edu.cn}}\quad \href{mailto:jfqu@ruc.edu.cn}{\nolinkurl{jfqu@ruc.edu.cn}}\\
\href{mailto:zhixuli@ruc.edu.cn}{\nolinkurl{zhixuli@ruc.edu.cn}}}
\newcommand{\papernotes}{$^{*}$: Equal contribution.\quad $^{\dagger}$: Corresponding authors.}
\newcommand{\paperkeywords}{information manipulation, multimodal conversational understanding, multimodal large language models}
\newcommand{\paperccs}{Computing methodologies~Natural language processing; Information systems~Multimedia information systems}

%% file: frontmatter.tex
% Same title / abstract layout as arXiv:2608.11201v1.
\thispagestyle{empty}

\noindent
\begin{minipage}[c]{0.55\linewidth}
  \raggedright
  \includegraphics[height=1.5cm]{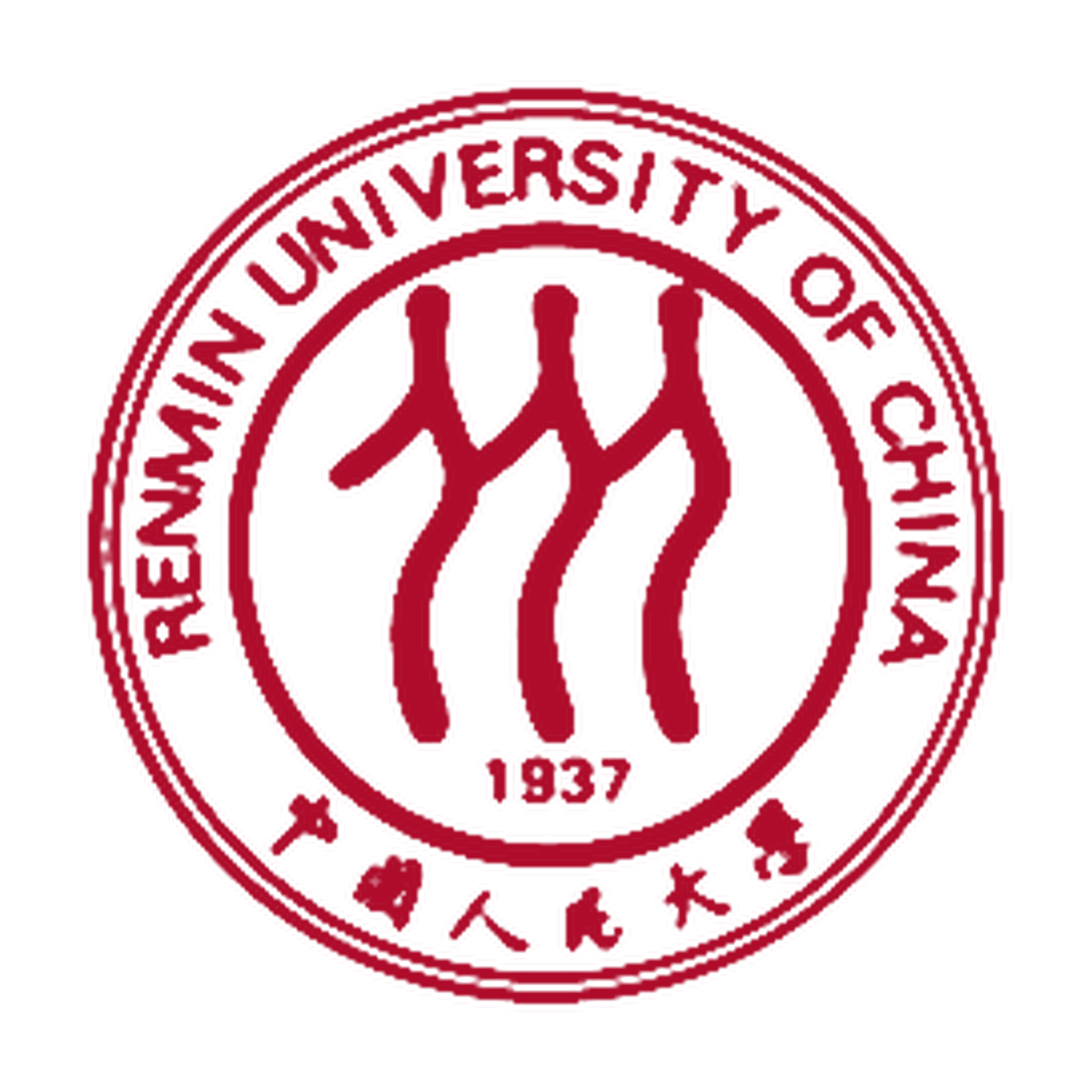}
\end{minipage}%
\begin{minipage}[c]{0.44\linewidth}
  \raggedleft
  {\rucsans\small\color{rucgray}Renmin University of China\par}

\end{minipage}\par

\vspace{0.35em}
\noindent{\color{rucline}\rule{\linewidth}{0.8pt}\par}
\vspace{1.0em}

\begin{center}
  {\ructitlefont\fontsize{13}{20}\selectfont\color{rucdark}
   \papertitle\par}
  \vspace{1.25em}
  {\normalsize\rmfamily\color{rucdark}\paperauthors\par}
  \vspace{0.22cm}
  {\footnotesize\rmfamily\color{rucgray}\paperaffiliations\par}
  {\footnotesize\rmfamily\itshape\color{rucgray}\papernotes\par}
  \vspace{0.12cm}
  {\scriptsize\rmfamily\color{rucgray}\paperemails\par}
\end{center}

\vspace{0.45em}
\begin{ructitlebox}
  \setlength{\parindent}{0cm}
  \setlength{\parskip}{0.14cm}
  \raggedright
  \nohyphens{\input{abstract}}
  \vspace{0.14cm}
  \vspace{-0.08cm}
\end{ructitlebox}

{\footnotesize\noindent\textbf{CCS Concepts:} \paperccs\par
\par}
\medskip

%% file: abstract.tex
Recent advances in MLLMs have improved social understanding, enabling models to recognize emotions and distinguish truthful from deceptive behaviors. However, existing benchmarks largely focus on coarse-grained emotion recognition or binary truth--deception detection, overlooking the diverse ways people manipulate information in conversations, such as withholding, leakage, fabrication, evasion, and defensive counterclaim. Understanding these behaviors requires reasoning about underlying facts, participant knowledge, and conversational context, making them difficult to evaluate reliably. To address this gap,  we introduce \textsc{InfoCon}, a multimodal benchmark for fine-grained \textbf{info}rmation manipulation in \textbf{con}versations. \textsc{InfoCon} identifies six fine-grained strategies and derives a coarse consistency label from their factual and interactional relations. We develop an evidence-grounded multi-agent annotation framework to construct and verify reliable labels from rich contextual evidence. 
We benchmark representative multimodal models and systematically analyze their fine-grained information manipulation understanding, providing a testbed beyond binary truth--deception judgments. Our benchmark dataset is available at the URL \url{https://huggingface.co/datasets/xilinghuiye/RoomReader-Full}.

%% file: body.tex
\begin{figure}[t]
    \centering
    \includegraphics[width=1\linewidth]{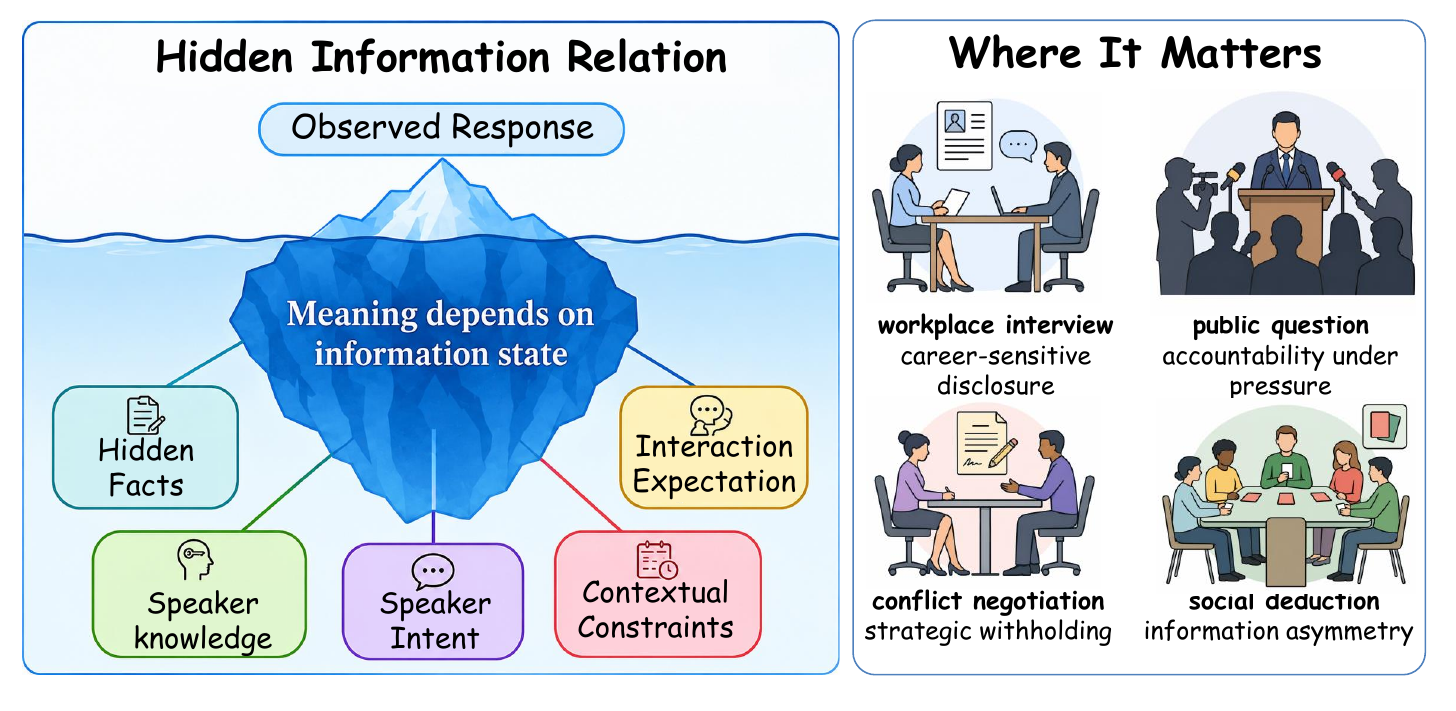}
\caption{Beneath the iceberg of everyday conversations lies a diverse landscape of information manipulation, which shapes how information is disclosed, withheld, leaked, fabricated, or avoided in different contexts.}
\Description{An iceberg links an observed conversational response to hidden facts, speaker knowledge, intent, contextual constraints, and interaction expectations. Example settings include negotiation, interviews, public questioning, and social deduction.}
\label{fig:motivation}
\end{figure}

\section{Introduction}
\label{sec:intro}

Understanding how people communicate and manage information is fundamental to multimodal intelligence. In real conversations, people may disclose facts, withhold sensitive details, leak concealed information, evade answers, or counter challenges. Multimodal large language models (MLLMs) jointly process visual, acoustic, and linguistic signals to understand human interactions. Multimodal emotion understanding has progressed from coarse affect recognition to fine-grained, context-aware reasoning~\cite{poria2019meld,cheng2024emotion,peng2026emotion}, while deception detection and social reasoning combine verbal and non-verbal cues to assess truthfulness~\cite{miah2025hidden,braverman2025decepbench,song2026thinkdeception}. These advances demonstrate MLLMs' growing ability to understand conversational behavior.

Yet existing research largely characterizes \emph{what a person feels} or \emph{whether a person is truthful}, overlooking \emph{how people manipulate and manage information}. As Figure~\ref{fig:motivation} illustrates, a response's meaning depends on hidden facts, speaker knowledge, and conversational expectations beyond surface audio-visual cues. Information manipulation arises in workplace interviews, public questioning, negotiation, and social interaction. Truthful communication need not imply full disclosure: speakers may withhold truth, leak concealed information, or evade questions without making false statements~\cite{mccornack1992information,mccornack2014information}; fabrication and defensive counterclaims can deliberately contradict underlying facts. These strategies have distinct communicative functions, but binary formulations collapse them. Recent multimodal deception benchmarks retain this coarse truth--deception focus, leaving finer information-management mechanisms largely unexplored~\cite{braverman2025decepbench,kang2025can}.

Advancing LLMs toward sophisticated affective intelligence therefore requires a benchmark of fine-grained information manipulation in realistic conversations, beyond coarse truthfulness judgments and across social circumstances. Construction requires more than inspecting isolated responses: annotators must establish underlying facts, speaker knowledge, expected disclosure, and the response's relation to the preceding interaction. Evidence for these relations may occur far beyond the target segment. Reliable ground-truth construction consequently requires rich context, whereas realistic, controlled model evaluation must restrict input to observable audio-visual interaction to measure multimodal understanding. Existing benchmarks do not explicitly address this tension between sufficient annotation evidence and restricted inference, hindering reliable supervision and controlled evaluation of these strategies.

To address this challenge, we introduce \textsc{InfoCon}, a multimodal benchmark for fine-grained \textbf{Info}rmation manipulation in \textbf{Con}versations. Table~\ref{tab:benchmark-comparison} compares its task, scale, granularity, and information-manipulation coverage with affect, deception, and social-interaction benchmarks. Models identify one of six strategies---\emph{active full disclosure}, \emph{selective withholding}, \emph{inadvertent leakage}, \emph{active fabrication}, \emph{evasive response}, and \emph{defensive counterclaim}---and derive its coarse consistency group. Our evidence-grounded multi-agent annotation framework jointly reconstructs facts, participant knowledge, interactional expectations, and target-response relations, followed by independent verification. Evaluation supplies only the target audio-visual clip: models infer the strategy from observable interaction without annotation-time context. This separation of \emph{context-rich ground-truth construction} from \emph{context-restricted inference} enables controlled evaluation of fine-grained information manipulation in multimodal conversations.

Our contributions are summarized as follows:
\begin{itemize}
    \item We formulate \textbf{fine-grained information manipulation understanding} in multimodal conversations as a social reasoning task, and introduce \textbf{\textsc{InfoCon}}, a benchmark with a six-way taxonomy that goes beyond conventional binary truth--deception judgments.

    \item We develop an \textbf{evidence-grounded multi-agent annotation framework} that incorporates underlying facts, participant knowledge, and interactional expectations to construct and verify reliable fine-grained supervision.
    
    \item We establish a \textbf{context-restricted evaluation protocol} and benchmark representative multimodal models, systematically revealing the challenges of fine-grained information manipulation understanding from short audio-visual conversational segments.
\end{itemize}

\providecommand{\infoconyes}{\textcolor{green!55!black}{\Large \ding{51}}}
\providecommand{\infoconno}{\textcolor{red!75!black}{\Large \ding{56}}}

\begin{table*}[t]
\centering
\caption{Comparison with representative affect, deception, and social-interaction benchmarks. \textsc{InfoCon} combines large-scale multimodal conversational clips with fine-grained information-strategy labels and context-rich gold annotation.}
\label{tab:benchmark-comparison}
\scriptsize
\setlength{\tabcolsep}{3.8pt}
\renewcommand{\arraystretch}{1.16}
\begin{adjustbox}{max width=\linewidth,max totalheight=0.80\textheight}
\begin{tabular}{@{}>{\raggedright\arraybackslash}p{2.35cm}>{\raggedright\arraybackslash}p{4.1cm}>{\centering\arraybackslash}p{1.3cm}>{\centering\arraybackslash}p{1.7cm}>{\centering\arraybackslash}p{2.0cm}>{\centering\arraybackslash}p{2.0cm}>{\centering\arraybackslash}p{2.4cm}@{}}
\toprule
\textbf{Benchmark} &
\textbf{Primary Task} &
\shortstack{\textbf{Dataset}\\\textbf{Size}} &
\shortstack{\textbf{Fine-grained}\\\textbf{Labels}} &
\shortstack{\textbf{Info-strategy}\\\textbf{Labels}} &
\shortstack{\textbf{Context-rich}\\\textbf{Gold}} &
\shortstack{\textbf{Information}\\\textbf{Management Focus}} \\
\midrule
MELD~\cite{poria2019meld} &
basic emotion recognition &
13,708 &
\infoconyes &
\infoconno &
\infoconno &
\infoconno \\

DFEW~\cite{jiang2020dfew} &
facial expression recognition &
16,372 &
\infoconyes &
\infoconno &
\infoconno &
\infoconno \\

MAFW~\cite{liu2022mafw} &
compound emotion recognition &
10,045 &
\infoconyes &
\infoconno &
\infoconno &
\infoconno \\

Real-life Trial~\cite{perez2015deception} &
binary deception detection &
121 &
\infoconno &
\infoconno &
\infoconno &
\infoconyes \\

Bag-of-Lies~\cite{gupta2019bag} &
binary deception detection &
325 &
\infoconno &
\infoconno &
\infoconno &
\infoconyes \\

DOLOS~\cite{guo2023audio} &
binary deception detection &
1,675 &
\infoconno &
\infoconno &
\infoconno &
\infoconyes \\

Box of Lies~\cite{soldner2019box} &
binary deception detection &
1,049 &
\infoconno &
\infoconno &
\infoconno &
\infoconyes \\

Werewolf Among Us~\cite{lai2023werewolf} &
social-deduction behavior &
26,647 &
\infoconyes &
\infoconno &
\infoconyes &
\infoconyes \\
\midrule
\textbf{\scalebox{1.2}{\textsc{InfoCon} (ours)}} &
\textbf{fine-grained information manipulation} &
\textbf{\textcolor{green!55!black}{\scalebox{1.2}{29,480}}} &
\infoconyes &
\infoconyes &
\infoconyes &
\infoconyes \\
\bottomrule
\end{tabular}
\end{adjustbox}
\vspace{2pt}
\begin{minipage}{0.98\textwidth}
\scriptsize
\end{minipage}
\end{table*}

\section{Related Work}
\label{sec:related}

\subsection{Multimodal Emotion Understanding}

Multimodal affective computing has progressed from coarse emotion recognition to nuanced affective understanding~\cite{busso2008iemocap,ghosal2019dialoguegcn,hu2021mmgcn,lian2023mer}. MELD~\cite{poria2019meld} and DFEW~\cite{jiang2020dfew} established large-scale conversational and in-the-wild video benchmarks; MAFW~\cite{liu2022mafw} added compound affective categories and multimodal cues. Recent MLLMs extend categorical recognition to instruction-following and affective reasoning~\cite{lian2025affectgpt,zhao2025r1,huang2026oneemo,huang2026nano}. Emotion-LLaMA~\cite{cheng2024emotion} integrates visual, auditory, and linguistic information for emotion recognition and reasoning; Emotion-LLaMAv2~\cite{peng2026emotion} further studies general-purpose reasoning about complex emotions. Beyond these substantial advances in human-affect understanding, information management remains underexplored. \textsc{InfoCon} benchmarks this distinct dimension through fine-grained information manipulation in multimodal conversations.

\subsection{Multimodal Deception Detection}

Deception detection traditionally distinguishes truthful from deceptive behavior~\cite{buller1996interpersonal,levine2014truth,perez2015experiments}. Information-manipulation theory also identifies omission, relevance manipulation, and deliberate ambiguity~\cite{mccornack1992information,mccornack2014information}; investigative-interview research emphasizes statement--evidence relations over universal behavioral cues~\cite{depaulo2003cues,vrij2008cognitive,vrij2019reading}. Multimodal studies combine visual, acoustic, and linguistic signals~\cite{newman2003lying,perez2014multimodal,perez2015verbal,9165161} to detect and interpret deception~\cite{miah2025hidden,braverman2025decepbench,kang2025can}. Werewolf Among Us~\cite{lai2023werewolf} supplies social-deduction scenarios with persuasion, strategic communication, and information asymmetry; ThinkDeception~\cite{song2026thinkdeception} pursues interpretable detection through progressive reasoning and reinforcement learning. Nevertheless, existing benchmarks and models largely retain binary truth--deception labels, collapsing disclosure, withholding, leakage, fabrication, evasion, and defensive counterclaim. \textsc{InfoCon} explicitly benchmarks these distinct information-manipulation strategies in multimodal conversations.

\subsection{Multimodal Social Reasoning}

Multimodal social reasoning increasingly addresses latent beliefs, intentions, and interaction dynamics. Social-IQ~\cite{zadeh2019social} evaluates multimodal social intelligence through question answering over natural social situations; MuMA-ToM~\cite{shi2025muma} focuses on multi-agent theory-of-mind reasoning involving beliefs and social goals. Social Genome~\cite{mathur2025social} grounds social reasoning in multimodal evidence, while SIV-Bench~\cite{kong2026siv} evaluates social scene understanding, social-state reasoning, and social-dynamics prediction. Complementing these advances beyond surface affect, \textsc{InfoCon} benchmarks how speakers manage information under asymmetric knowledge and conversational expectations, advancing fine-grained, information-aware social reasoning.

\section{\textsc{InfoCon} Task and Taxonomy}
\label{sec:task}

\subsection{Task Definition}

\textsc{InfoCon} formulates fine-grained information manipulation understanding as a classification task over multimodal conversational clips. Given an audio-visual interaction clip
\begin{equation}
    x_i = (v_i, a_i),
    \label{eq:input}
\end{equation}
where $v_i$ and $a_i$ denote the visual and audio streams, respectively, the goal is to identify how the speaker manages information in the observed interaction.

The model-visible input contains only the target audio-visual clip, without the extended contextual information used during annotation. The task therefore evaluates how much of the context-grounded information strategy can be recovered from locally observable multimodal cues. Gold labels describe the strategy established by source evidence; predictions estimate that strategy from the restricted observation. Source transcripts, role truth, and annotation rationales belong to construction evidence and must not be passed to evaluated models. Distinct source contexts can produce similar local clips, so reliable gold annotation does not imply that every instance is uniquely resolvable from the clip. This information asymmetry is the evaluation condition shared by all models, rather than an assumption that intention or speaker knowledge is directly observable. Let the fine-grained label space be
\begin{equation}
    \mathcal{Y}
    =
    \{D, W, L, F, E, C\},
    \label{eq:fine-label-space}
\end{equation}
where $D$, $W$, $L$, $F$, $E$, and $C$ denote \emph{active full disclosure}, \emph{selective withholding}, \emph{inadvertent leakage}, \emph{active fabrication}, \emph{evasive response}, and \emph{defensive counterclaim}, respectively.

Given the model-visible input $x_i$, a model $f_{\theta}$ predicts the fine-grained information-manipulation strategy as
\begin{equation}
    \hat{y}_i
    =
    \arg\max_{y \in \mathcal{Y}}
    p_{\theta}(y \mid x_i).
    \label{eq:fine-prediction}
\end{equation}

The six fine-grained categories are further grouped according to whether the speaker's observed behavior is consistent with the behavior expected under truthful communication~\cite{levine2014truth,mccornack1992information}. This grouping provides a coarse-grained consistency label in addition to the primary six-way prediction.

\begin{figure*}[t]
\centering
  \includegraphics[width=\linewidth]{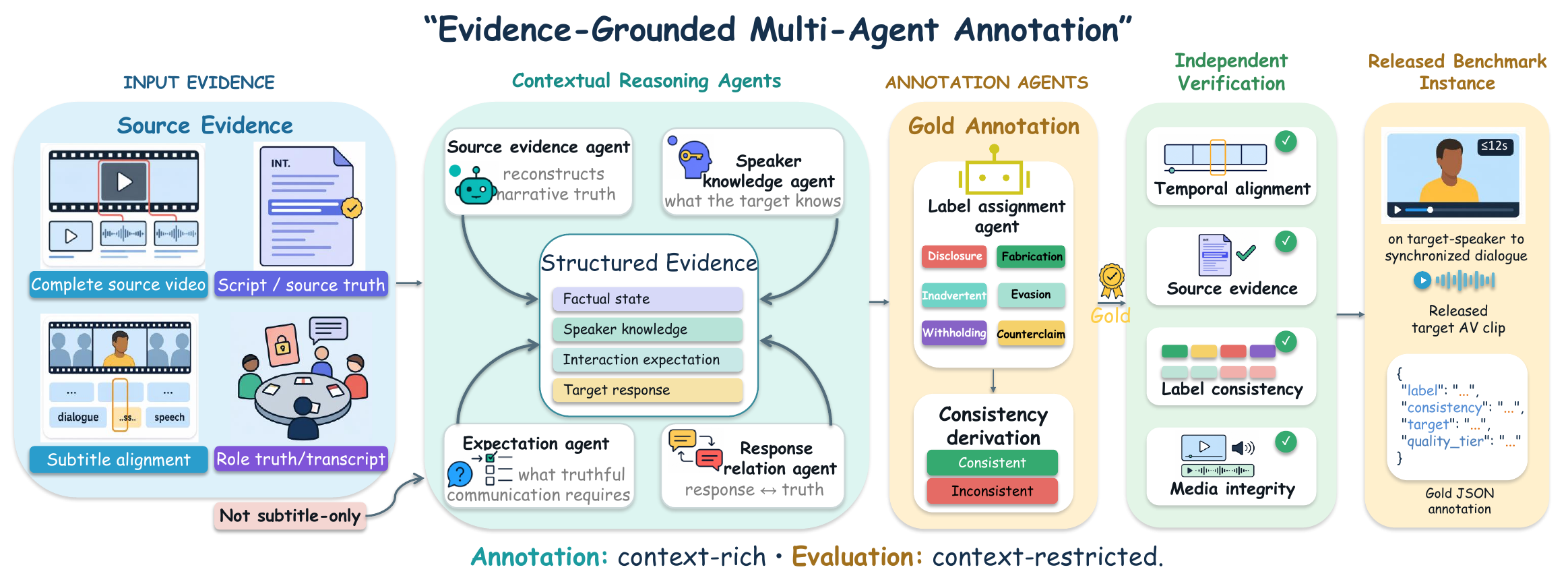}
\caption{Overview of the \textsc{InfoCon} benchmark construction protocol. Rich source-grounded contextual evidence is used to construct and verify fine-grained labels, while only the target audio-visual clips are released for model evaluation.}
\Description{A schematic showing source media, contextual evidence modeling, fine-grained annotation, independent verification, and the final released audio-visual clips.}
\label{fig:construction}
\end{figure*}

\subsection{Information-Manipulation Taxonomy}
\label{taxonomy}

The six information-manipulation strategies are organized into two groups according to their consistency with the behavior expected under truthful communication~\cite{mccornack1992information,mccornack2014information}. The first group contains behaviors that remain consistent with truthful communication, although the extent or manner of information disclosure differs. The second group contains behaviors that deviate from truthful communication through fabrication, avoidance, or defensive contradiction~\cite{buller1996interpersonal}.

Formally, the six-way label space is partitioned into
\begin{equation}
    \mathcal{Y}_{\mathrm{con}}
    =
    \{D, W, L\},
    \qquad
    \mathcal{Y}_{\mathrm{inc}}
    =
    \{F, E, C\},
    \label{eq:coarse-partition}
\end{equation}
where $\mathcal{Y}_{\mathrm{con}}$ represents behaviors consistent with truthful communication and $\mathcal{Y}_{\mathrm{inc}}$ represents behaviors inconsistent with it. Here, \emph{consistency} is the benchmark's operational grouping of factual compatibility and response--expectation relations, not a claim that a response is complete or non-misleading. Withholding may be factually compatible yet pragmatically misleading; evasion can violate the expected response without stating a falsehood. The coarse label is always obtained by the fixed mapping in Equation~\ref{eq:coarse-partition}.

\paragraph{Consistent behaviors.}
These behaviors do not explicitly contradict the grounded facts, but differ in disclosure completeness and intentionality.

\begin{itemize}
    \item \textbf{Active Full Disclosure ($D$):} The speaker voluntarily provides all relevant information truthfully, without concealment or omission.

    \item \textbf{Selective Withholding ($W$):} The speaker intentionally withholds part of the relevant information, such as negative or sensitive details, while disclosing only selected information.

    \item \textbf{Inadvertent Leakage ($L$):} The speaker unintentionally reveals information that was not intended to be disclosed, such as through a spontaneous remark or an unintended slip.
\end{itemize}

\paragraph{Inconsistent behaviors.}
These behaviors deviate from truthful communication through fabrication, avoidance, or defensive contradiction.

\begin{itemize}
    \item \textbf{Active Fabrication ($F$):} The speaker deliberately provides information that contradicts the underlying facts or the truth.

    \item \textbf{Evasive Response ($E$):} The speaker avoids directly answering the request by deflecting the question, changing the topic, or providing a vague or non-committal response.

    \item \textbf{Defensive Counterclaim ($C$):} After being challenged or questioned, the speaker forcefully denies or counters the claim, potentially reversing or contradicting the underlying facts rather than directly addressing the issue.
\end{itemize}

The taxonomy distinguishes information-manipulation strategies that may be conflated under a binary truth--deception formulation~\cite{mccornack2014information}. In particular, \emph{selective withholding} differs from \emph{evasive response} because the former involves selective disclosure of truthful information, whereas the latter avoids directly addressing the request. Similarly, \emph{inadvertent leakage} differs from \emph{active full disclosure} in the intentionality of disclosure, while \emph{defensive counterclaim} differs from \emph{active fabrication} in its reactive nature. These distinctions form the basis of the fine-grained evaluation in InfoCon. Operationally, the relevant disclosure is bounded by the local exchange: $W$ supplies part of the requested information while omitting a known material fact, whereas $E$ fails to fill the requested information slot. For example, stating a known transaction while concealing its sensitive item is $W$; deflecting the request for that item is $E$. A false account is $F$, while a denial or reversal organized as a response to a challenge is $C$; falsity alone does not distinguish the pair. Likewise, $D$ and $L$ both reveal facts, but the evidence must distinguish deliberate disclosure from an unintended revelation. These boundaries are interpreted through the structured facts, knowledge, expectation, and response relation, rather than inferred from demeanor alone.

\subsection{Consistency-Level Evaluation}

Although the primary task is six-way information-manipulation classification, \textsc{InfoCon} also provides a coarse-grained evaluation of whether the observed behavior is consistent with truthful communication. The consistency label is deterministically derived from the fine-grained prediction:
\begin{equation}
    \hat{c}_i
    =
    \phi(\hat{y}_i)
    =
    \begin{cases}
        \mathrm{Consistent}, & \hat{y}_i \in \mathcal{Y}_{\mathrm{con}},\\
        \mathrm{Inconsistent}, & \hat{y}_i \in \mathcal{Y}_{\mathrm{inc}}.
    \end{cases}
    \label{eq:coarse-prediction}
\end{equation}

Thus, the two evaluation levels can be summarized as
\begin{equation}
    x_i
    \xrightarrow{\;f_{\theta}\;}
    \hat{y}_i
    \xrightarrow{\;\phi\;}
    \hat{c}_i,
    \quad
    \hat{y}_i \in \mathcal{Y},
    \quad
    \hat{c}_i \in
    \{\mathrm{Consistent},\mathrm{Inconsistent}\}.
    \label{eq:two-level-task}
\end{equation}

The six-way classification constitutes the primary task of InfoCon, while the derived consistency label provides a coarse-grained evaluation interface that facilitates comparison with conventional truth--deception-oriented models.

% Match Figure 3's visible height to the fully typeset Table 2.
\newsavebox{\infoconStatisticsBox}
\newlength{\infoconOverviewHeight}
\begin{figure*}[t]
\centering
\sbox{\infoconStatisticsBox}{%
\begin{minipage}{0.58\textwidth}
  \centering
  \scriptsize
  \setlength{\tabcolsep}{3pt}
  \renewcommand{\arraystretch}{1.12}

  \resizebox{\linewidth}{!}{%
    \begin{adjustbox}{max width=\linewidth,max totalheight=0.80\textheight}
\begin{tabular}{@{}lccc@{}}
    \toprule
    \textbf{Statistic} & \textbf{Overall} & \textbf{Scripted film} & \textbf{Werewolf} \\
    \midrule
    \multicolumn{4}{@{}l}{\textbf{Dataset and split}} \\
    Total clips
      & 29,480
      & 18,237
      & 11,243 \\
    Training / test clips
      & 25,458 / 4,022
      & 15,714 / 2,523
      & 9,744 / 1,499 \\
    Source units
      & 166
      & 15
      & 151 \\
    \midrule
    \multicolumn{4}{@{}l}{\textbf{Consistency}} \\
    Consistent
      & 17,591 (59.7\%)
      & 14,120 (77.4\%)
      & 3,471 (30.9\%) \\
    Inconsistent
      & 11,889 (40.3\%)
      & 4,117 (22.6\%)
      & 7,772 (69.1\%) \\
    \midrule
    \multicolumn{4}{@{}l}{\textbf{Fine-grained information strategies}} \\
    Active full disclosure
      & 11,843 (40.2\%)
      & 10,749 (59.0\%)
      & 1,094 (9.7\%) \\
    Selective withholding
      & 5,362 (18.2\%)
      & 3,117 (17.1\%)
      & 2,245 (20.0\%) \\
    Inadvertent leakage
      & 386 (1.3\%)
      & 254 (1.4\%)
      & 132 (1.2\%) \\
    Active fabrication
      & 2,969 (10.1\%)
      & 2,086 (11.4\%)
      & 883 (7.8\%) \\
    Evasive response
      & 5,337 (18.1\%)
      & 665 (3.6\%)
      & 4,672 (41.6\%) \\
    Defensive counterclaim
      & 3,583 (12.2\%)
      & 1,366 (7.5\%)
      & 2,217 (19.7\%) \\
    \bottomrule
    \end{tabular}
\end{adjustbox}%
  }
\end{minipage}%
}
\setlength{\infoconOverviewHeight}{%
  \dimexpr\ht\infoconStatisticsBox+\dp\infoconStatisticsBox\relax}
\begin{minipage}[t]{0.405\textwidth}
  \vspace{0pt}
  \centering
  % Trim only the PDF's white border and preserve the chart's aspect ratio.
  \includegraphics[height=\infoconOverviewHeight,
    trim=14bp 14bp 14bp 14bp,clip]{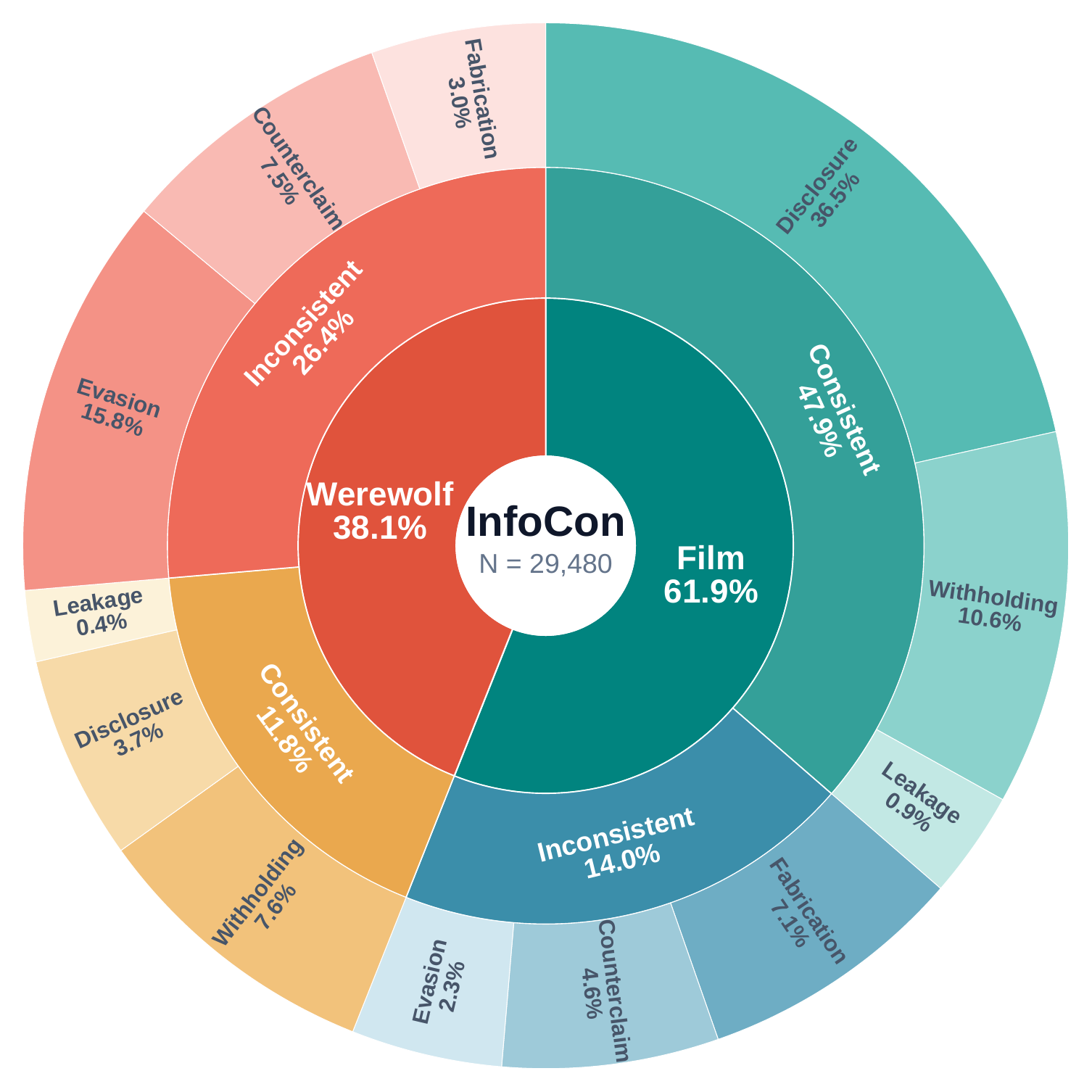}\par
\end{minipage}%
\hfill
\begin{minipage}[t]{0.58\textwidth}
  \vspace{0pt}
  \centering
  \raisebox{\dp\infoconStatisticsBox}{\usebox{\infoconStatisticsBox}}\par
\end{minipage}

\par\nointerlineskip
\begin{minipage}[t]{0.405\textwidth}
  \vspace{0pt}
  \captionsetup{position=bottom,skip=6pt}
\caption{Distribution of \textsc{InfoCon} instances by source, consistency, and strategy.}
  \Description{A sunburst partitions 29,480 clips into film and Werewolf sources, their coarse consistency groups, and six strategies. The adjacent table lists the corresponding counts, proportions, and train/test split sizes.}
\label{fig:infocon-overview}
\end{minipage}%
\hfill
\begin{minipage}[t]{0.58\textwidth}
  \vspace{0pt}
  \captionsetup{position=bottom,skip=6pt}
\captionof{table}{Detailed statistics of the \textsc{InfoCon} dataset. Percentages are computed within each column. The official dataset contains 29,480 clips from scripted films and social-deduction games.}
  \label{tab:infocon-statistics}
\end{minipage}
\end{figure*}

\section{Benchmark Construction and Protocol}
\label{sec:construction}
Figure~\ref{fig:construction} summarizes source collection, contextual reconstruction, label assignment, and independent verification.

\subsection{Data Sources}

\textsc{InfoCon} combines scripted film/video with social-deduction games~\cite{lai2023werewolf}: narrative-grounded facts and authored exchanges complement role-grounded knowledge asymmetry, strategic questioning, and confrontation.

Film/video sources require both the complete video and an independent source-of-truth text, such as a screenplay, stage play, or literary source. Subtitles align dialogue temporally; they are not the sole ground-truth evidence.

The sources cover the same information relations under different interactional pressures. Their label distributions illustrate this complementarity (Table~\ref{tab:infocon-statistics}): full disclosure accounts for 59.0\% of film clips but 9.7\% of game clips, while evasion accounts for 3.6\% and 41.6\%, respectively. Together they cover fact-compatible disclosure and adversarial response management.

We identify responses grounded in explicit information relations, localize them into audio-visual clips of at most 12 seconds, and apply the shared six-way taxonomy (Section~\ref{taxonomy}).

\subsection{Evidence-Grounded Multi-Agent Annotation}

Gold construction separates \emph{contextual reconstruction}, \emph{label assignment}, and \emph{independent verification}\cite{zhang2026generative}. The same utterance can have different strategies depending on facts, speaker knowledge, and interactional expectations; isolated surface cues do not establish these relations.

Each evidence bundle includes the complete source video, independent narrative or role truth, aligned subtitles/transcripts, and speaker/role information where available. This recovers evidence before or after the response while preserving temporal alignment.

Four contextual agents reconstruct complementary relations\cite{liu2026omniverifier}. The \emph{source-evidence agent} establishes facts from the narrative or interaction history; the \emph{speaker-knowledge agent} determines what the speaker knows at that moment; the \emph{expectation agent} determines what truthful communication requires; and the \emph{response-relation agent} relates the response to those facts and expectations.

Their outputs are consolidated into a structured evidence representation\cite{zhang2026omniverifier,liu2026vidforensics}
\begin{equation}
    \mathcal{E}_i =
    \left(
        F_i,\,
        K_i,\,
        X_i,\,
        R_i
    \right),
    \label{eq:structured-evidence}
\end{equation}
Here, $F_i$, $K_i$, $X_i$, and $R_i$ denote factual state, speaker knowledge, interactional expectation, and response relation, respectively.

Given $\mathcal{E}_i$, a dedicated label-assignment agent selects $y_i\in\mathcal{Y}$ from the six strategies in Section~\ref{taxonomy}. This decision uses factual truth, speaker knowledge, expectation, and response relation, rather than surface wording or demeanor alone. The coarse label is derived, not independently predicted:
\begin{equation}
    c_i=\phi(y_i),\qquad y_i\in\mathcal{Y},
    \label{eq:gold-consistency}
\end{equation}
using the same partition as Equation~\ref{eq:coarse-partition}.

Independent verification checks (1) \emph{temporal alignment} with the target response, (2) \emph{source evidence} supporting facts and knowledge, (3) \emph{label consistency} with the taxonomy, and (4) \emph{media integrity}, including synchronization and target-speaker centering. Failed instances are rejected or revised before admission.

\begin{figure*}
    \centering
    \includegraphics[width=1\linewidth]{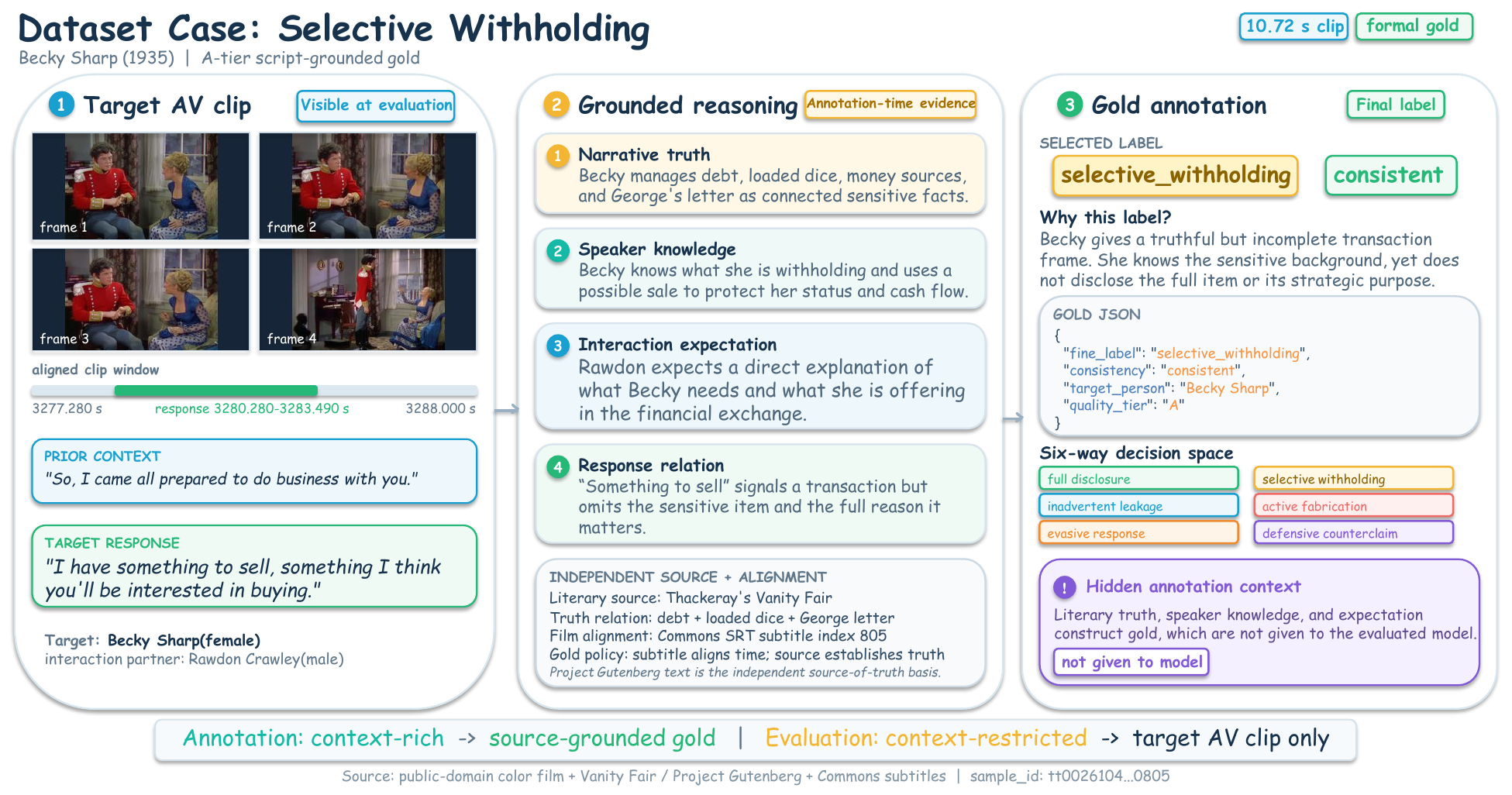}
\caption{An annotation case of selective withholding, illustrating context-rich gold construction from source-grounded evidence and context-restricted evaluation on the target audio-visual clip.}
    \Description{A 10.72-second Becky Sharp clip is labeled selective withholding. The response indicates something for sale but omits its sensitive content. Narrative facts, speaker knowledge, and interaction expectations support the gold annotation and remain hidden during model evaluation.}
\label{fig:case}
\end{figure*}

The complete annotation pipeline can therefore be summarized as
\begin{equation}
    \mathcal{S}_i
    \xrightarrow{\mathcal{A}_{\mathrm{ctx}}}
    \mathcal{E}_i
    \xrightarrow{\mathcal{A}_{\mathrm{label}}}
    (y_i,c_i)
    \xrightarrow{\mathrm{Verify}}
    \mathcal{B}_i,
    \label{eq:annotation-pipeline}
\end{equation}
Here, $\mathcal{S}_i$ is the source evidence bundle, $\mathcal{A}_{\mathrm{ctx}}$ the contextual agents, and $\mathcal{B}_i$ an accepted instance. Its gold annotation stores the fine label, derived consistency, target identity, and evidence reference; model input contains only the localized synchronized clip.

Thus, \emph{annotation is context-rich and source-grounded, whereas evaluation is context-restricted}: $\mathcal{E}_i$ supports reliable labels without supplying privileged evidence to evaluated models.

\definecolor{RRHead}{HTML}{EEF1F5}
\definecolor{RRGrid}{HTML}{CAD1DA}
\definecolor{RRBlueBand}{HTML}{DDEBFF}
\definecolor{RRGreenBand}{HTML}{E2F2E7}
\definecolor{RROrangeBand}{HTML}{FFE9C7}
\definecolor{RRWinner}{HTML}{E9E9E9}
\definecolor{RRBestFill}{HTML}{E7EDFF}
\definecolor{RRSecondFill}{HTML}{E4F4E8}
\definecolor{RRThirdFill}{HTML}{FFF3BF}
\definecolor{RRBestText}{HTML}{2452C8}
\definecolor{RRSecondText}{HTML}{258243}
\definecolor{RRThirdText}{HTML}{BC9818}
\definecolor{RRDeltaText}{HTML}{C52222}
\definecolor{RRGainFill}{HTML}{F5F6F8}
\definecolor{RRModText}{HTML}{4C5665}

\newcolumntype{Y}{>{\centering\arraybackslash}X}
\newcommand{\rrfirst}[1]{\textcolor{RRBestText}{\textbf{#1}}}
\newcommand{\rrsecond}[1]{\textcolor{RRSecondText}{\textbf{#1}}}
\newcommand{\rrthird}[1]{\textcolor{RRThirdText}{\textbf{#1}}}
\newcommand{\rrgain}[1]{\textcolor{RRDeltaText}{\scriptsize\(\uparrow\)#1}}
\newcommand{\rrmod}[1]{\textcolor{RRModText}{\textsf{\footnotesize #1}}}
\newcommand{\rrdash}{\textemdash}
\newcommand{\rrgroup}[2]{%
  \addlinespace[1.4pt]
  \rowcolor{#1}\multicolumn{10}{@{}l}{\textsf{\footnotesize\bfseries\hspace{0.45em}#2}}\\[-0.4pt]
}

\begin{table*}[t]
\centering
\caption{Main results on the \textsc{InfoCon} Core-Test split ($N=4{,}022$). Macro-F1 is the primary metric. \rrfirst{Blue}, \rrsecond{green}, and \rrthird{yellow} numbers mark the \rrfirst{top-1}, \rrsecond{top-2}, and \rrthird{top-3} scores in each column across all available results. \rrgain{Red gain} rows report absolute improvements over the corresponding zero-shot base model. Training regimes are separated explicitly.}
\label{tab:main-results}
\scriptsize
\setlength{\tabcolsep}{2.75pt}
\renewcommand{\arraystretch}{1.14}
\arrayrulecolor{RRGrid}
\resizebox{\textwidth}{!}{%
\begin{adjustbox}{max width=\linewidth,max totalheight=0.80\textheight}
\begin{tabular}{@{}>{\raggedright\arraybackslash}p{3.85cm}c*{8}{c}@{}}
\toprule[0.95pt]
\rowcolor{RRHead}
\textbf{Model} & \textbf{Mod.} & \multicolumn{4}{c}{\textbf{Six-way \textsc{InfoCon}}} & \multicolumn{4}{c}{\textbf{Coarse Consistency}} \\
\rowcolor{RRHead}
 & & \textbf{Macro-F1} & \textbf{Weighted-F1} & \textbf{Recall} & \textbf{Acc.} & \textbf{Macro-F1} & \textbf{Weighted-F1} & \textbf{Recall} & \textbf{Acc.} \\
\midrule
\rrgroup{RROrangeBand}{Conventional audio-visual models}
HiCMAE~\cite{sun2024hicmae} & \rrmod{V+A} &
0.1975 & 0.3171 & 0.2515 & 0.4110 &
0.6889 & 0.6958 & 0.6872 & 0.6992 \\

CA2ST~\cite{11224949} & \rrmod{V+A} &
0.1796 & 0.2960 & 0.2529 & \rrthird{0.4212} &
\rrthird{0.6929} & \rrthird{0.7000} & \rrthird{0.6911} & \rrthird{0.7039} \\

CAV-MAE~\cite{gong2023cavmae} & \rrmod{V+A} &
0.2351 & \rrthird{0.3317} & 0.2586 & 0.3864 &
0.6564 & 0.6645 & 0.6554 & 0.6691 \\

\midrule
\rrgroup{RRBlueBand}{General-purpose MLLMs}

Qwen3-VL-8B-Instruct~\cite{bai2025qwen3} & \rrmod{V} &
0.1168 & 0.1606 & 0.1520 & 0.2051 &
0.4545 & 0.4580 & 0.4565 & 0.4560 \\

InternVL3.5-8B~\cite{wang2025internvl3} & \rrmod{V} &
0.0582 & 0.0736 & 0.1668 & 0.2106 &
0.3595 & 0.4034 & 0.5000 & 0.5612 \\

Qwen2-Audio-7B-Instruct~\cite{chu2024qwen2} & \rrmod{A} &
0.0577 & 0.0728 & 0.1647 & 0.2079 &
0.3582 & 0.4020 & 0.4953 & 0.5559 \\

Doubao-Seed-2.0-lite-260428~\cite{bytedanceseed2026seed2} & \rrmod{V+A} &
0.2124 & 0.2599 & 0.2206 & 0.2663 &
0.5453 & 0.5526 & 0.5446 & 0.5525 \\

\rowcolor{RRBlueBand!25}Qwen3-Omni-30B-A3B-Instruct~\cite{xu2025qwen3} & \rrmod{V+A} & 0.2355 & 0.2577 & 0.2482 & 0.2578 & 0.5661 & 0.5797 & 0.5717 & 0.5943 \\

\midrule
\rrgroup{RRGreenBand}{Emotion-specialized MLLMs}
Emotion-LLaMA-v2~\cite{peng2026emotion} & \rrmod{V+A} & 0.0707 & 0.1143 & 0.1657 & 0.2681 & 0.4096 & 0.5547 & 0.5007 & 0.6807 \\
\rowcolor{RRGreenBand!27}AffectGPT~\cite{lian2025affectgpt} & \rrmod{V+A} & \rrthird{0.2386} & 0.2702 & \rrthird{0.2691} & 0.3138 & 0.4957 & 0.4826 & 0.5532 & 0.5192 \\
OneEmo~\cite{huang2026oneemo} & \rrmod{V+A} & 0.1475 & 0.1813 & 0.2007 & 0.2218 & 0.5530 & 0.5663 & 0.5645 & 0.5879 \\
Nano-EmoX~\cite{huang2026nano} & \rrmod{V+A} & 0.0969 & 0.1327 & 0.1679 & 0.2383 & 0.4878 & 0.5103 & 0.5233 & 0.5601 \\
R1-Omni~\cite{zhao2025r1} & \rrmod{V+A} & 0.0846 & 0.0998 & 0.1635 & 0.1548 & 0.4383 & 0.4332 & 0.5006 & 0.4911 \\
Emotion-LLaMA~\cite{cheng2024emotion} & \rrmod{V+A} & 0.0720 & 0.0894 & 0.1637 & 0.2120 & 0.4328 & 0.4659 & 0.5102 & 0.5608 \\

\midrule
\rrgroup{RRWinner}{\textsc{InfoCon}-adapted MLLMs}
\rowcolor{RRWinner}\textbf{Qwen3-Omni + \textsc{InfoCon}} & \rrmod{V+A} & \rrfirst{0.3931} & \rrfirst{0.4856} & \rrfirst{0.4152} & \rrfirst{0.5266} & \rrfirst{0.7202} & \rrfirst{0.7266} & \rrfirst{0.7179} & \rrfirst{0.7300} \\
\rowcolor{RRGainFill}\quad\textcolor{RRDeltaText}{\scriptsize gain vs. base} & & \rrgain{0.1576} & \rrgain{0.2279} & \rrgain{0.1670} & \rrgain{0.2688} & \rrgain{0.1541} & \rrgain{0.1469} & \rrgain{0.1462} & \rrgain{0.1357} \\
\rowcolor{RRWinner}\textbf{Emotion-LLaMA + \textsc{InfoCon}} & \rrmod{V+A} & \rrsecond{0.2736} & \rrsecond{0.3658} & \rrsecond{0.3395} & \rrsecond{0.4580} & \rrsecond{0.6997} & \rrsecond{0.7071} & \rrsecond{0.6976} & \rrsecond{0.7121} \\
\rowcolor{RRGainFill}\quad\textcolor{RRDeltaText}{\scriptsize gain vs. base} & & \rrgain{0.2016} & \rrgain{0.2764} & \rrgain{0.1758} & \rrgain{0.2460} & \rrgain{0.2669} & \rrgain{0.2412} & \rrgain{0.1874} & \rrgain{0.1513} \\
\bottomrule[0.95pt]
\end{tabular}
\end{adjustbox}%
}
\arrayrulecolor{black}
\vspace{2pt}
\begin{minipage}{0.96\textwidth}
\end{minipage}
\end{table*}

\subsection{Dataset Statistics, Splits, and Analysis}

As is illustrated in Figure~\ref{fig:infocon-overview} and Table~\ref{tab:infocon-statistics}, the resulting \textsc{InfoCon} dataset contains 29,480 instances, including 18,237 film/video instances and 11,243 social-deduction game instances. The official dataset is divided into training and test partitions, containing 25,458 and 4,022 instances, respectively. The 166 source units comprise 15 films and 151 game units (Table~\ref{tab:infocon-statistics}). The six labels are imbalanced: leakage has 386 instances overall (1.3\%), including 199 in the official training partition (0.78\%) and 187 in Core-Test (4.65\%; Table~\ref{tab:classwise-results}). We therefore interpret six-way Macro-F1 together with per-class scores, since accuracy and weighted F1 can conceal minority-class failures.

To reduce source-level leakage, we split the dataset at the source level rather than at the individual clip level. All clips originating from the same complete film or game unit are assigned to the same partition, preventing highly similar interactions from appearing across training and test sets.

\subsection{Annotation Case Study}

Figure~\ref{fig:case} presents a representative case of \emph{selective withholding}. In the 10.72-second clip from \textit{Becky Sharp} (1935), Becky tells Rawdon, ``I have something to sell, something I think you'll be interested in buying.'' Viewed alone, the response suggests a transaction but does not reveal whether relevant information is being concealed.

The annotation-time evidence resolves this ambiguity by reconstructing four grounded relations: the narrative truth, Becky's knowledge of the sensitive information, Rawdon's expectation of a direct explanation, and the relation between the response and the underlying facts. Together, these show that Becky truthfully signals a transaction while intentionally omitting the sensitive item and its strategic relevance. The instance is therefore labeled \emph{selective withholding} and mapped to the \emph{consistent} group.

The gold label is grounded in the independent literary source, while subtitles are used only for temporal alignment. These contextual cues are used only during annotation; evaluation is performed on the localized target audio-visual clip alone. This case illustrates why a gold label can be supported by the full source even when a local prediction remains ambiguous: the model must use the observable response and its multimodal delivery without being given the withheld item.

\section{Experiments}

\subsection{Experimental Setup}
\label{sec:experimental-setup}

\paragraph{Data split and protocol.}
We use the official source-level split of \textsc{InfoCon}, containing
25,458 training instances and 4,022 Core-Test instances. Clips from the
same film or social-deduction game are assigned to the same split to
reduce source-level leakage. For trainable baselines, a source-disjoint
validation subset is held out from the training data for checkpoint
selection, while Core-Test is used only for final evaluation. Following
the context-restricted setting, models receive only the target clip of
at most 12 seconds, without the contextual evidence used to construct
the gold labels.

\paragraph{Compared models.}
We evaluate three groups of baselines: conventional audio-visual
models, general-purpose MLLMs with visual, audio, or audio-visual
inputs, and emotion-specialized multimodal models. Conventional models
are trained on the \textsc{InfoCon} training split, whereas unadapted
MLLMs are evaluated using a unified zero-shot instruction. We further
report \textsc{InfoCon}-adapted variants using LoRA~\cite{hu2021lora} or supervised
fine-tuning.

\paragraph{Metrics.}
We evaluate both six-way information-manipulation classification and
the derived binary consistency task. For both tasks, we report
Macro-F1, weighted F1, recall, and accuracy. Six-way Macro-F1 is the primary score because it gives each strategy equal weight despite class imbalance; class-wise precision, recall, and F1 expose which strategies account for the aggregate result.

\begin{figure}
    \centering
    \includegraphics[width=0.5\linewidth]{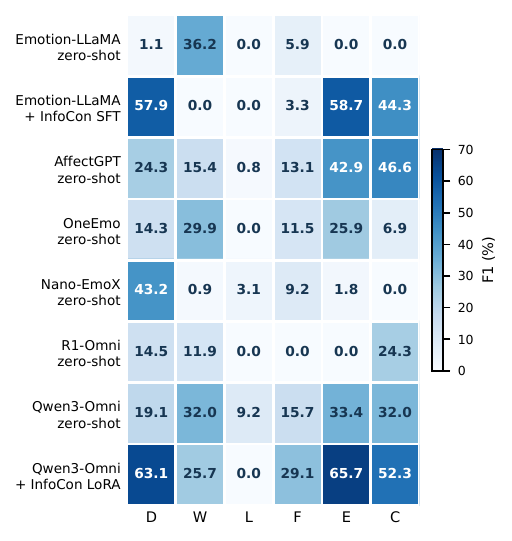}
    \caption{Class-wise performance across multimodal models on the \textsc{InfoCon} Core-Test split. Each cell reports F1 (\%); columns follow the $D/W/L/F/E/C$ order in Section~\ref{taxonomy}.}
    \Description{A heatmap reports per-class F1 for eight base or adapted multimodal systems. Adapted Qwen3-Omni scores 63.1, 25.7, 0.0, 29.1, 65.7, and 52.3 percent across disclosure, withholding, leakage, fabrication, evasion, and counterclaim.}
\label{fig:classwise-performance}
\end{figure}

\subsection{Main Results}

Table~\ref{tab:main-results} shows that fine-grained information
manipulation remains challenging for existing models: the best
zero-shot MLLM, AffectGPT, achieves only 0.2386 six-way Macro-F1. Interestingly,
the conventional audio-visual model CAV-MAE (0.2351) performs
comparably to substantially larger multimodal LLMs, suggesting that
general multimodal capacity alone does not resolve this task and
that fine-grained interaction reasoning remains a key bottleneck.
This also highlights the difficulty of recovering subtle information
states from short context-restricted audio-visual clips. These rows compare the stated training and modality conditions: CAV-MAE uses \textsc{InfoCon} supervision, whereas unadapted MLLMs are zero-shot. The adaptation pairs isolate the observed benefit relative to each base model; comparisons between different model families do not isolate a modality or adaptation-method effect.

In contrast, coarse consistency scores are substantially higher; for
example, CA2ST obtains 0.6929 coarse Macro-F1 but only 0.1796 on the
six-way task. This gap indicates that existing models can often
distinguish broadly consistent from inconsistent behavior, but still
struggle to identify the specific information-manipulation strategy.

\begin{table*}[htbp]
  \centering
  \caption{Performance comparison of Qwen3-Omni on the Real-life Trial benchmark under Zero-shot and Supervised Fine-Tuning (SFT) settings.}
  \label{tab:reallife_trial_results}
  \begin{adjustbox}{max width=\linewidth,max totalheight=0.80\textheight}
\begin{tabular}{l l c c c}
    \toprule
    \textbf{Benchmark} & \textbf{Setting} & \textbf{Accuracy (\%)} & \textbf{Macro-F1 (\%)} & \textbf{$\Delta$ Macro-F1} \\
    \midrule
    \rowcolor{RRGainFill}\multicolumn{5}{l}{\textit{Zero-shot Evaluation}} \\
    \multirow{2}{*}{Real-life Trial} 
      & Qwen3-Omni zero-shot & 52.89 & 44.76 & -- \\
      & \textsc{InfoCon}-trained zero-shot & \textbf{61.98} & \textbf{59.31} & \textcolor{red}{+14.55} \\
    \midrule
    \rowcolor{RRGainFill}\multicolumn{5}{l}{\textit{Supervised Fine-Tuning (SFT) Evaluation}} \\
    \multirow{2}{*}{Real-life Trial} 
      & External-only SFT & 63.64 & 62.27 & -- \\
      & \textsc{InfoCon} $\rightarrow$ External SFT & \textbf{68.60} & \textbf{68.54} & \textcolor{red}{+6.27} \\
    \bottomrule
  \end{tabular}
\end{adjustbox}
\end{table*}

Finally, \textsc{InfoCon} adaptation yields clear improvements, increasing
Qwen3-Omni from 0.2355 to 0.3931 and Emotion-LLaMA from 0.0720 to
0.2736 in six-way Macro-F1. The gains are also consistently reflected
in weighted F1, recall, accuracy, and coarse consistency performance.
Importantly, improvements are observed for both a general-purpose
multimodal LLM and an emotion-specialized model, indicating that the
benefit of \textsc{InfoCon} supervision is not tied to a particular model
family. Table~\ref{tab:main-results} reports point estimates; small score differences should be read as descriptive rankings rather than evidence of statistical significance.

Figure~\ref{fig:classwise-performance} further reveals substantial class-wise
variation across models. Among the six information-manipulation
strategies, inadvertent leakage is consistently the most challenging,
with several models obtaining near-zero F1, while selective withholding
also remains difficult for many systems. In contrast, active full
disclosure, evasive response, and defensive counterclaim are more
effectively captured by the stronger models. This uneven performance
suggests that fine-grained information-manipulation understanding is a
more demanding task that requires models to capture subtle differences
in speaker knowledge, disclosure completeness, intentionality, and
interactional behavior.

\subsection{Benchmark Quality}

To validate annotation reliability, we evaluate both human agreement
and model-backbone consistency. Motivated by prior work demonstrating
the potential of LLMs for annotation tasks~\cite{gilardi2023chatgpt},
\textsc{InfoCon} annotations denote the final dataset labels produced by
our evidence-grounded multi-agent annotation framework instantiated with
GPT~\cite{openai2026gpt5.5} and accepted after source-evidence and
label-consistency verification; they are not single raw model predictions.
Human annotations are independently provided by two graduate students
with prior experience in data-oriented research. They re-annotate 100
stratified samples using the same contextual evidence available during
dataset construction, while being blind to the original \textsc{InfoCon}
annotations and to each other's decisions.
As shown in Table~\ref{tab:annotation-validation}, we report agreement
between Human and \textsc{InfoCon} annotations and inter-human agreement. The human audit uses the full evidence packet and therefore assesses label reproducibility, rather than human performance under clip-only inference. Its 100-sample size also means that the agreement estimates have finite-sample uncertainty.
We additionally instantiate the same evidence-grounded multi-agent
annotation framework with Claude~\cite{anthropic2026claude} and GPT,
while keeping the evidence, taxonomy, and annotation instructions
unchanged, to evaluate consistency across different underlying models.
We report accuracy/agreement, Macro-F1, and Cohen's
$\kappa$~\cite{cohen1960coefficient}, where $\kappa$ measures agreement
beyond chance. As a common rule of thumb, $\kappa>0.60$ indicates
substantial agreement, while $\kappa>0.80$ indicates almost perfect
agreement.

\begin{table}[t]
\centering
\caption{Human and multi-model validation of \textsc{InfoCon} annotations.
}
\label{tab:annotation-validation}
\small
\setlength{\tabcolsep}{4pt}

\resizebox{\columnwidth}{!}{%
\begin{adjustbox}{max width=\linewidth,max totalheight=0.80\textheight}
\begin{tabular}{lccc}
\toprule
\textbf{Setting} &
\textbf{Acc./Agreement} &
\textbf{Macro-F1} &
\textbf{$\kappa$} \\
\midrule

Human vs.\ \textsc{InfoCon} Annotation
& \textcolor{black}{0.860}
& \textcolor{black}{0.842}
& \textcolor{black}{0.819} \\

Human 1 vs.\ Human 2
& \textcolor{black}{0.830}
& \textcolor{black}{0.806}
& \textcolor{black}{0.782} \\

Claude vs.\ GPT
& \textcolor{black}{0.894}
& \textcolor{black}{0.826}
& \textcolor{black}{0.851} \\

\bottomrule
\end{tabular}
\end{adjustbox}%
}
\end{table}

\definecolor{gain}{RGB}{35,145,90}
\definecolor{loss}{RGB}{190,70,70}
\definecolor{softgray}{RGB}{105,105,105}

\begin{table*}[htbp]
    \centering
    \caption{Class-wise performance of Qwen3-Omni on the \textsc{InfoCon}
    Core-Test split before and after \textsc{InfoCon} adaptation.
    Precision (P), recall (R), and F1 are reported in percentages.
    $\Delta$ F1 denotes the absolute change in F1 after adaptation.}
    \label{tab:classwise-results}
    \small
    \setlength{\tabcolsep}{8pt}
    \begin{adjustbox}{max width=\linewidth,max totalheight=0.80\textheight}
\begin{tabular}{l r c c r}
        \toprule
        \textbf{Category} &
        \textbf{Support} &
        \textbf{Zero-shot P/R/F1} &
        \textbf{\textsc{InfoCon}-adapted P/R/F1} &
        \textbf{$\Delta$ F1} \\
        \midrule
        \multicolumn{5}{l}{
            \textcolor{softgray}{\textit{Consistent strategies}}
        } \\[-2pt]
        Active Full Disclosure
            & 1,224 & 36.1 / 13.0 / 19.1 & 48.4 / 90.6 / 63.1 & \textcolor{gain}{\bfseries $\uparrow$44.0} \\
        Selective Withholding
            & 846   & 24.2 / 46.9 / 32.0 & 60.5 / 16.3 / 25.7 & \textcolor{loss}{\bfseries $\downarrow$6.3} \\
        Inadvertent Leakage
            & 187   & 5.8 / 21.9 / 9.2   & 0.0 / 0.0 / 0.0    & \textcolor{loss}{\bfseries $\downarrow$9.2} \\
        \addlinespace[3pt]
        \multicolumn{5}{l}{
            \textcolor{softgray}{\textit{Inconsistent strategies}}
        } \\[-2pt]
        Active Fabrication
            & 388   & 20.2 / 12.9 / 15.7 & 29.5 / 28.6 / 29.1 & \textcolor{gain}{\bfseries $\uparrow$13.3} \\
        Evasive Response
            & 972   & 37.5 / 30.0 / 33.4 & 86.7 / 52.9 / 65.7 & \textcolor{gain}{\bfseries $\uparrow$32.3} \\
        Defensive Counterclaim
            & 405   & 47.1 / 24.2 / 32.0 & 45.9 / 60.7 / 52.3 & \textcolor{gain}{\bfseries $\uparrow$20.3} \\
        \bottomrule
    \end{tabular}
\end{adjustbox}
\end{table*}

\subsection{Transferability to Out-of-Domain Data}

We evaluate transfer to Real-life Trial~\cite{perez2015deception}, a widely used benchmark of 121 truthful/deceptive courtroom videos with a different domain, distribution, and binary annotation scheme. ``Zero-shot'' refers to the target domain: the base model has no \textsc{InfoCon} adaptation, while the \textsc{InfoCon}-trained model first learns the source six-way task and then performs the target binary task without target-domain training. External-only SFT uses target supervision; \textsc{InfoCon} $\rightarrow$ External SFT uses source supervision followed by target supervision.

Table~\ref{tab:reallife_trial_results} shows substantial gains without target-domain training: zero-shot Macro-F1 rises from 44.76 to 59.31 (+14.55 points), and accuracy from 52.89\% to 61.98\%. This cross-dataset gain indicates that \textsc{InfoCon} captures transferable information-manipulation and interactional cues rather than dataset-specific label patterns. With target supervision, source initialization improves Macro-F1 from 62.27 to 68.54 and accuracy from 63.64\% to 68.60\%. These results demonstrate a strong transferable representation for downstream deception understanding that complements task-specific supervision.

Transfer from six-way strategies to external binary deception highlights the broader value of fine-grained, evidence-grounded annotations. It suggests generalizable relations among factual consistency, speaker behavior, and response strategies. Strong target-zero-shot gains further indicate potential for low-resource or few-shot deception detection, where high-quality annotations are difficult to collect.

\subsection{Fine-Grained Difficulty Analysis and Future Directions}
\label{sec:fine-grained-analysis}

To better understand the effect of \textsc{InfoCon} supervision, we
report class-wise precision, recall, and F1 of Qwen3-Omni before and
after adaptation in Table~\ref{tab:classwise-results}.
The gains are clearly non-uniform across categories. The largest
improvements appear for \emph{active full disclosure},
\emph{evasive response}, and \emph{defensive counterclaim}, with F1
increases of 44.0, 32.3, and 20.3 points, respectively. Active
fabrication also improves by 13.3 points, showing that \textsc{InfoCon}
supervision enables the model to learn substantially stronger
fine-grained information-management patterns.
In contrast, \emph{selective withholding} and \emph{inadvertent
leakage} remain difficult after adaptation. In particular, withholding
suffers from low recall, while the minority leakage category is not
successfully recovered. Since selective withholding contains more
examples than several categories that improve substantially, the
difficulty cannot be attributed to class imbalance alone. These
categories require finer reasoning about speaker knowledge,
intentionality, and the boundary between omission and revelation.
Future work may therefore explore class-balanced training, targeted
hard-example supervision, and explicit modeling of speaker knowledge
and disclosure intentionality to further improve these challenging
cases. Leakage remains part of the primary six-way task: excluding it would hide a core intentionality distinction and its minority-class failure. Nonzero leakage F1 before adaptation (9.2\% for Qwen3-Omni) also shows that its post-adaptation failure does not establish that the category is intrinsically unlearnable.

\paragraph{Interpreting ambiguity and source diversity.}
A source-grounded label and a uniquely identifiable clip-level strategy are different properties. Missing context can leave $D/L$, $W/E$, or $F/C$ difficult to distinguish even when the full evidence supports a clear gold label; these are taxonomy-based ambiguity pairs, not measured confusion frequencies. Scripted acting and strategic gameplay provide complementary interaction settings with different incentives and label distributions. These differences motivate source-aware interpretation of aggregate scores and further evaluation of clip-level ambiguity. The Real-life Trial transfer results demonstrate useful supervision across domains and label spaces; the 121-video target set is one external domain and does not by itself quantify variation across all deployment settings.To mitigate the ambiguity caused by missing clip-level context, future work could incorporate long-horizon dialogue history modeling or cross-clip memory mechanisms, enabling the model to infer speaker intent within a more complete interactive trajectory. Furthermore, extending cross-domain evaluation beyond courtroom settings to broader real-world interactions—such as negotiations and everyday conversations—remains an important direction for future work.

\section{Conclusion}

We presented \textsc{InfoCon}, a benchmark for fine-grained social-cue
understanding under information asymmetry. Unlike conventional approaches
that primarily focus on binary deception detection or coarse-grained
affective understanding, \textsc{InfoCon} characterizes social responses
through a six-way taxonomy that distinguishes different strategies of
information disclosure, concealment, leakage, fabrication, avoidance, and
defensive counterclaim. 
To construct reliable fine-grained supervision, we
developed an evidence-grounded multi-agent annotation framework that models
the underlying facts, contextual relations, and expected conditions of an
interaction before assigning and verifying the final social-cue labels.
Importantly, while rich contextual evidence is used during annotation,
models are evaluated solely on the target audio-visual clips, preserving
the information asymmetry between annotation and inference. 
Through comprehensive experiments and qualitative analyses, we investigate the
ability of existing multimodal models to recognize these fine-grained
social cues and demonstrate the value of \textsc{InfoCon} for training and
evaluating social understanding models. We hope \textsc{InfoCon} can serve as a challenging and reliable testbed for future
research on multimodal social reasoning and information-aware interaction
understanding.

\section*{Ethical Considerations}

\textsc{InfoCon} is intended as a research benchmark for studying multimodal social reasoning and fine-grained information manipulation in conversational settings. The labels in the benchmark describe context-dependent communicative behaviors rather than stable personal traits, and model predictions should not be interpreted as definitive judgments about an individual's honesty, intent, or character.

\bibliographystyle{ACM-Reference-Format}
\bibliography{references}